\documentclass[letterpaper,conference]{ieeeconf}

\IEEEoverridecommandlockouts                              

\usepackage{graphicx}
\usepackage{amssymb}  
\usepackage{hyperref,tcolorbox}
\usepackage{subcaption}
\usepackage{enumerate}
\usepackage{cite}
\usepackage[keeplastbox]{flushend}
\usepackage{amsmath,bm}
\usepackage[algo2e]{algorithm2e} 

\newcommand\norm[1]{\left\lVert#1\right\rVert}

\usepackage{scrextend}

\title{\LARGE \bf
High-Speed High-Accuracy Spatial Curve Tracking\\ Using Motion Primitives in Industrial Robots 
}

\def\dq{\delta q}
\def\qdot{\dot q}
\def\qddot{\ddot q}

\def\nustar{\nu^*}

\def\ez{e_z}

\def\sq{^2}

\author{Honglu He\thanks{Electrical, Computer, and Systems Engineering, Rensselaer Polytechnic Institute, {\tt heh6@rpi.edu luc6@rpi.edu weny2@rpi.edu}},
        Chen-lung Lu,
        Yunshi Wen,
Glenn Saunders\thanks{Manufacturing Innovations Center, Rensselaer Polytechnic Institute, {\tt saundg@rpi.edu}
}, Pinghai Yang\thanks{GE Research, US, {\tt Pinghai.Yang@ge.com schoonov@ge.com}
}, Jeffrey Schoonover,\\
Agung Julius\thanks{Electrical, Computer, and Systems Engineering, Rensselaer Polytechnic Institute, {\tt agung@ecse.rpi.edu wenj@rpi.edu}
}, John T.~Wen
}

\begin{document}
\maketitle
\thispagestyle{empty}
\pagestyle{empty}

\begin{abstract}
Industrial robots are increasingly deployed in applications requiring an end effector tool to closely track a specified path, such as in spraying and welding.
Performance and productivity present possibly conflicting objectives: tracking accuracy, path speed, and motion uniformity. 
Industrial robots are programmed through motion primitives consisting of waypoints connected by pre-defined motion segments, with specified parameters such as path speed and blending zone. 
The actual executed robot motion depends on the robot joint servo controller and joint motion constraints (velocity, acceleration, etc.) which are largely unknown to the users.
Programming a robot to achieve the desired performance today is time-consuming and mostly manual, requiring tuning a large number of coupled parameters in the motion primitives.  The performance also depends on the choice of additional parameters: possible redundant degrees of freedom, location of the target curve, and the robot configuration.
This paper presents a systematic approach to optimize the robot  motion primitives for performance.  The approach first selects the static parameters, then the motion primitives, and finally iteratively update the waypoints to minimize the tracking error. 
The ultimate performance objective is to maximize the path speed subject to the tracking accuracy and speed uniformity constraints over the entire path.  
We have demonstrated the effectiveness of this approach in simulation for ABB and FANUC robots for two challenging example curves, and experimentally for an ABB robot.  Comparing with the baseline using the current industry practice, the optimized performance shows over 200\% performance improvement.

\end{abstract}
\par \textbf{\textit{Keywords-}}
\textbf{Industrial Robot, Motion Primitive, Path Optimization, Redundancy Resolution, Trajectory Tracking}

\section{INTRODUCTION}
\label{sec:intro}
Industrial robots are increasingly deployed in applications such as spray coating \cite{spray_coating}, arc welding \cite{arc_welding}, deep rolling \cite{deep_rolling}, surface grinding \cite{surface_grinding}, cold spraying \cite{cold_spray}, etc., where the tool center point (TCP) frame attached to the end effector needs to track complex geometric paths in both position and orientation. In most applications, the task performance is characterized by the motion speed (how long it takes to complete the task), motion uniformity (how much the speed varies along the path), and motion accuracy (the maximum position and orientation tracking errors along the path). 
There are generally two ways to program industrial robots for the motion tracking task: (1) The motion primitive method that uses  vendor-specific proprietary robot programming languages (e.g., \cite{Inform_manual,RAPID_manual,KAREL_manual,Vplus_manual,VAL3_manual,KRL_manual}) consisting of a sequence of pre-defined motion primitives, and (2) The streaming setpoint method which uses an external command mode to stream the desired joint position as setpoints to the low level joint servo controller (e.g., \cite{MotoPlus,EGM,LLI,Kuka_open}). 
The motion primitive method is far more widely used in industries.  It  decomposes the desired motion with predefined motion primitives consisting of waypoints connected by motion segments.  The streaming setpoint method is not limited to a small set of motion primitives, but it typically results in poorer tracking accuracy due to the lower sampling rate 
and the additional filtering and latency. 

To program an industrial robot to follow a complex path using motion primitives, the most common approach is to use it like a machine tool \cite{Robodk_sim}, focusing only on the TCP motion. There are several advanced offline programming software (e.g., \cite{Robodk,RobotMaster,Octobpuz}) that convert a dense set of TCP waypoint 
to the robot program of a given robot vendor. 
The actual robot motion from the execution of the robot program will depend on the motion primitives and their parameters (waypoint locations, commanded path speed along the motion segment, size of the blending zone between motion segments) which are affected by the robot joint servo controller and the robot joint motion constraints (joint velocity and acceleration limits).  Robot controller and robot joint constraints are largely unknown to the user.  Therefore, programming a robot to achieve the desired performance is currently a time-consuming and largely manual exercise in tuning the motion primitive parameters and the result may be far from optimal.  Compounding the challenges are the impact of additional parameters: a redundant
roll degree of freedom in the tool frame, relative pose between
the target curve and the robot base, and the robot configuration
(out of the 8 possible choices in a 6 degrees of freedom (DoF) revolute robot).

The streaming setpoint method avoids motion blending, but the actual robot motion is difficult to predict, again because of the unknown low level controller and robot motion constraints, in addition to the latency and filtering present in the streaming operation. 
Motion planning tools such as Tesseract \cite{Tesseract} (based on TrajOpt \cite{Trajopt}), tries to optimize the robot joint path for path speed subject to tracking error and speed uniformity constraints.  Such optimization involves a large number of variables and is plagued by computation time, presence of local minima, and the lack of accurate prediction of the actual robot motion.



\begin{figure}[h!]
    \centering
    \includegraphics[width=0.30\textwidth]{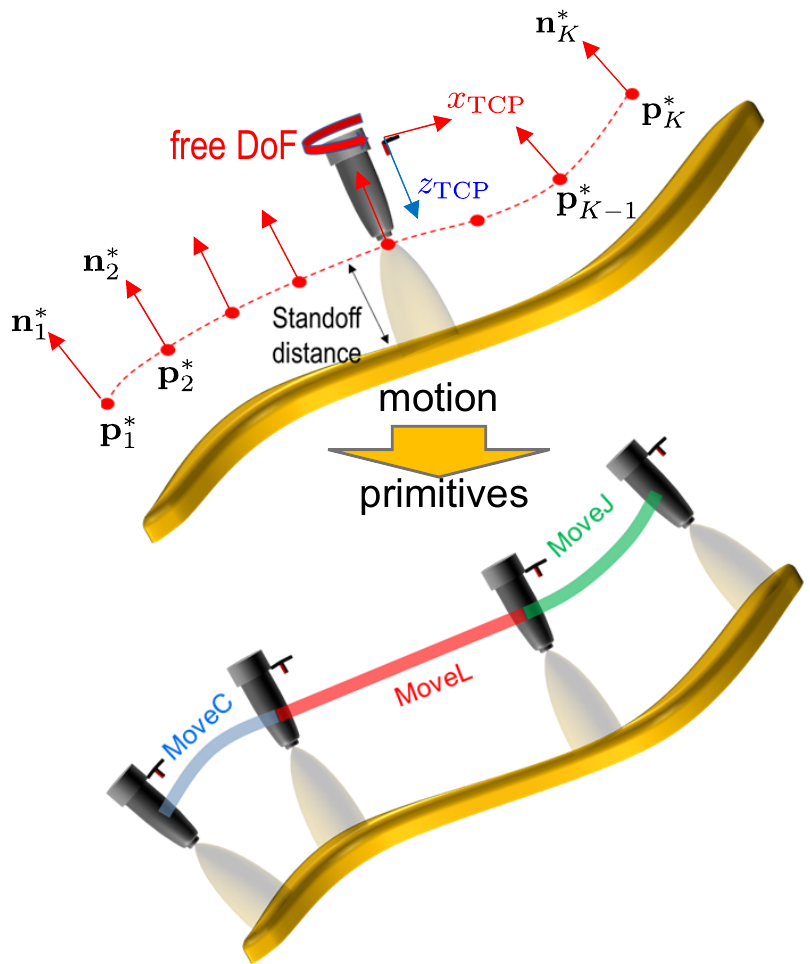}
    \caption{Illustration of the motion primitive method for path following.}
    \label{fig:f1}
\end{figure}

A key motivation of this work is the \emph{cold spray} application which impinges particles onto a surface at high speed to form protective coating \cite{cold_spray}.  For such applications, the speed of the TCP needs to be \emph{high and uniform} to guarantee fast execution and uniform distribution of the deposited material.  
As shown in Fig.~\ref{fig:f1}, the spraying path is given as a dense sequence of points $\{\mathbf{p}^\ast_i\}_{i=1}^K$ for the TCP position and the spraying direction in the negative tool $z$-axis is given by the surface normal $\{\mathbf{n}^\ast_i\}_{i=1}^K$.   Note that these data specify 5-DoF of the TCP at each point, with the remaining free DoF being the rotation about the tool $z$-axis.  Once the TCP frame is fixed, there are multiple corresponding robot joint configurations (inverse kinematics solutions).  
We also have an additional 6 free DoF in the position and orientation of the curve, parameterized by the position of the center of mass $\mathbf{p}_{curve}$ and the angle-product representation of the orientation $\bm{\beta}_{curve}$.  
The motion primitive method first chooses all these free parameters, we call this step {\em redundancy resolution}.  The TCP frames are then converted to motion primitives as shown in Fig.~\ref{fig:f1} for execution by the robot controller. 
%
%
This paper presents a new approach to optimizing the motion primitives by decomposing the problem into a series of subproblems. We first use a combination of evolution algorithm and local optimization to resolve redundancy. Then we use a greedy algorithm for fitting the curve with the least number of motion primitives. Finally, an iterative gradient descent algorithm updates the waypoints to minimize the worst case tracking error.  To evaluate this approach, the industry co-authors of this work suggested two test curves, curve one contains a range of high curvature moves and curve two is the leading edge of a sample turbine fan blade.  A baseline approach is developed based on the current industry practice.  The evaluation metric is the average path speed subject to the specified trajectory tracking and path speed uniformity constraints.  We implemented the motion primitive optimization approach on an ABB IRB 6640 robot in simulation (using RobotStudio) and physical testbed, and on an FANUC M710iC-70 robot in simulation (using RoboGuide).  In all cases, the performance vastly improves over the baseline, ranging over 200\% to 300\%.  

\noindent \textbf{Notation:} Throughout the paper, $\mathbf{p}\in \mathbb{R}^3$ denotes the Cartesian position of the TCP, $(v,\omega)$ denotes the linear and angular velocity of TCP, $p_0 
\in \mathbb{R}^3$ denote the TCP position at the robot zero configuration, $\bm{\beta} \in \mathbb{R}^3$ denotes the angle-product representation of the tool frame orientation,  
$\mathbf{q}\in \mathbb{R}^6$ denotes the joint position of the robot. The function $\mathtt{fwd}:\mathbf{q}\rightarrow(\mathbf{p},\bm{\beta})$ denotes the forward kinematic map. We also use the notations $\mathbf{p}(\mathbf{q})$ and $\bm{\beta}(\mathbf{q})$ to represent the forward kinematics. The notation $e_z(\mathbf{q}) \in \mathbb{R}^3$ is a unit vector denoting the $z$-axis of the TCP in the robot base frame. 
The TCP spatial velocity is denoted by $\begin{bmatrix} \omega \\ v \end{bmatrix} = J(\mathbf{q}) \dot{\mathbf{q}} $, where $J$ is the Jacobian matrix.



\begin{tcolorbox}\textbf{Contribution:} We present a method for calculating a set of motion primitives and their respective parameters (see Sec. \ref{sec:primitives} for more details) so that the TCP tracks the desired positions and orientations with sufficient accuracy and speed uniformity (see Sec. \ref{sec:spec} for more details) as fast as possible.
\end{tcolorbox}

\section{Motion Primitives and Their Executions}
\label{sec:primitives}

\subsection{Motion Primitives}
Generally all robot programming languages support three types of motion primitives: MoveL, MoveC and MoveJ, which are described below. 
\begin{itemize}
    \item MoveL:  The TCP moves linearly in Cartesian space and linearly in angle-product orientation from current pose to desired pose. This command is parameterized by the initial and final TCP positions and angle-product orientations.
    \item MoveC:  The TCP moves on a circular arc in Cartesian space and linearly in angle-product orientation from current pose to desired pose. This command is parameterized by the initial and final TCP positions, an intermediate TCP position (to define the circular arc), and the initial and final angle-product orientations, an intermediate orientation (to define rotation orientation).
    \item MoveJ:  The robot moves linearly in the \emph{joint angles space}. Thus, the TCP path may not be linearly nor circular. This command is parametrized by the initial and final joint angles.
\end{itemize}


\subsection{Blending}
Given a series of motion primitive segments, the desired tool path is  blended together to avoid sharp turns \cite{6-5_spline}. The size of the blending zone is specified with the primitives. Larger blending zone means smoother transition to next segment, and hence higher and more uniform speed, but at the cost of large tracking error at the waypoint (intersection of motion segments). Small blending zone improves the tracking accuracy but could lead to sharp corner requiring slower speed with larger variation to meet the motion constraints. In current practice, expert human operators perform extensive manual adjustments of the waypoints, blending regions, and motion segment commanded speeds to ensure that tracking accuracy and cycle time requirements are both met.  In our approach, we find the smallest uniform blending zone that avoids excessive speed variation around waypoints to balance between the two objectives. 

\section{Specifications and Proposed Approach}
\label{sec:spec}

Based on the cold spray application, the trajectory tracking performance specification is given by: \\
\noindent \textbf{1. Positional tracking accuracy:} the maximum positional tracking error of the TCP is less than 0.5 mm.\\
\textbf{2. Orientation tracking accuracy:} the maximum orientation tracking error of tool $z$-axis is less than 3$^\circ$.\\
\textbf{3. Speed uniformity:} the standard deviation of the TCP speed is less than 5\%.\\

\begin{figure}[h]
    \centering
    \includegraphics[width=0.48\textwidth]{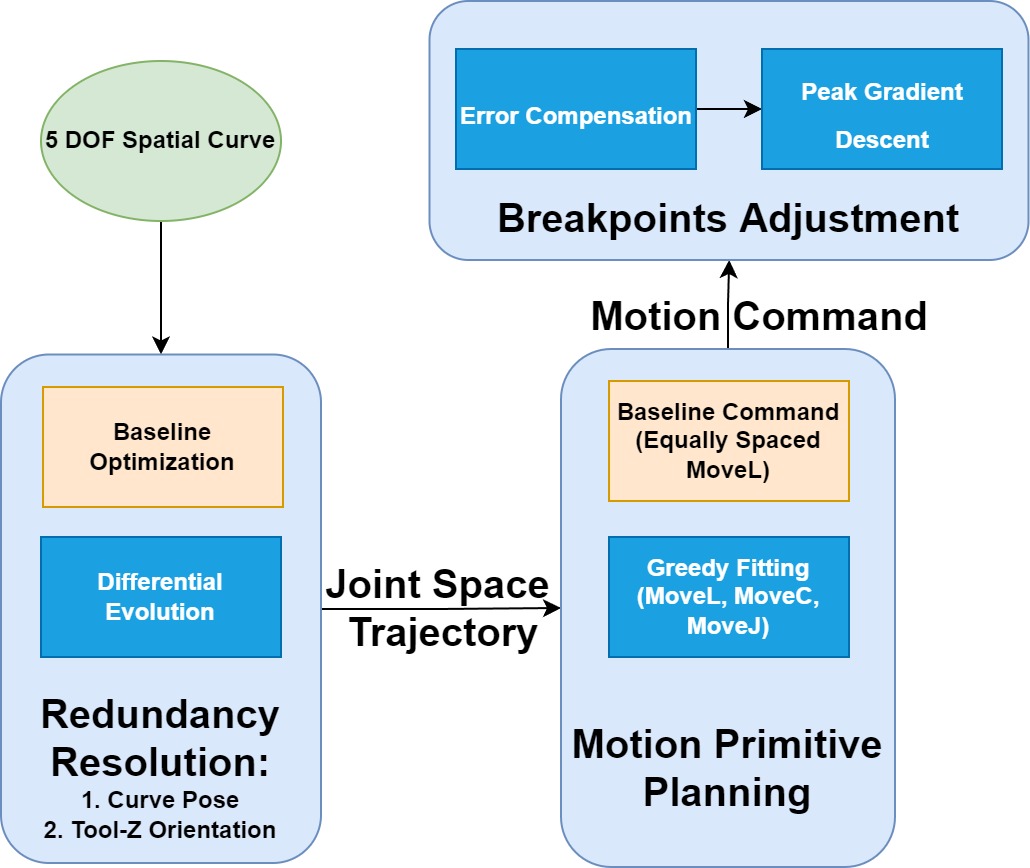}
    \caption{Proposed Workflow for Spatial Curve Tracking with Motion Primitives.
    }
    \label{fig:flowchart}
\end{figure}

Our approach decomposes the motion primitive optimization problem into three steps as shown in Fig. \ref{fig:flowchart}. \\
{\bf 1. Redundancy Resolution}: Optimize the redundant DoF (curve pose: ($\mathbf{p}_{curve}$,$\bm{\beta}_{curve}$)), the redundant DoF along the path, and the robot arm configuration. The output of this process is the full 6 DoF reference path for the TCP and the corresponding joint path. \\
{\bf 2. Motion Primitive Planning}: Generate a sequence of motion primitives (waypoints, blending zones, and path speed) to track the reference path. \\
{\bf 3. Waypoint Adjustment}: The motion sequence is executed in simulations or experiments.  Tracking error from the executions is used to optimize the motion primitives parameters. \\
These steps are described in greater details below.

\subsection{Redundancy Resolution}
In contrast to computer numerical control (CNC) machines, robot pose significantly affects the TCP motion due to the robot joint constraints, including the joint position, velocity, and acceleration limits.  There are multiple redundant DoF that may be exploited to place the robot in a more advantageous configuration without compromising the TCP motion:
\begin{itemize}
    \item Tool orientation: Rotation around the tool $z$-axis at each trajectory point.
    \item Curve pose: Position $\mathbf{p}_{curve}$ and orientation $\bm{\beta}_{curve}$ (6-doF).
    \item Robot pose: One out of maximum eight joint configurations due to the non-uniqueness of the inverse kinematic map.
\end{itemize}
Our approach is to first resolve the tool redundant DoF by finding the complete robot joint path based on a given curve pose and robot pose.  At each point on this joint path, we can find the path velocity limit based on the joint velocity and acceleration limits.  We then choose the curve pose and robot pose to maximize the lowest path velocity limit.  The joint acceleration limit is due to the motor torque constraint.  Since the arm inertia is configuration dependent, the joint acceleration limit is also configuration dependent (hence, an outstretched arm has a lower acceleration limit).  We will show how to estimate this limit through simulation.

\subsubsection{Tool Orientation Resolution}
\label{sec:toolorientation}

For a given part position and orientation, $(\mathbf{p}_{curve},\bm{\beta}_{curve})$, and robot pose, we have 
$\{\mathbf{p}^\ast_i, \mathbf{n}^\ast_i\}_{i=1}^K$ in the robot base frame and the initial joint angle $\mathbf{q}_0$. 
We select the redundant DoF so that the TCP motion remains on the path with minimal incremental joint motion.  Since joint motion and TCP motion are related through the arm Jacobain $J(\mathbf{q})$, this procedure tends to steer the robot away from singularities without compromising TCP tracking.  We pose the incremental optimization as a quadratic program (QP):
Given $\mathbf{q}_{k-1}$, $k=1,\ldots,K$, find the incremental motion $\dq$ to move to the next point $(\mathbf{p}^\ast_k, \mathbf{n}^\ast_k)$ without excessive joint motion and within the joint limits:
\begin{equation}
\begin{aligned}
    &\min_{\dq} \norm{J_r(\mathbf{q}_{k-1}) \dq -\nustar}\sq + W_q\norm{\dq}\sq \\
    &\mbox{subject to } q_{\min} \preccurlyeq \mathbf{q}_{k-1}+\dq \preccurlyeq q_{\max}
\end{aligned}
\label{eq:QP_reduced}
\end{equation}
where $(q_{min},q_{max})$ are joint limits, $W_q$ is a weightfactor,  $\nustar$ is the TCP position and surface normal increment, and $J_r$ is the reduced Jacobian mapping $\dq$ to the position and surface normal increment: 
\begin{equation}
    \nustar := \begin{bmatrix} \mathbf{\ez}(\mathbf{q}_{k-1})-\mathbf{n}^*_{k} \\ \mathbf{p}(\mathbf{q}_{k-1})-\mathbf{p}^*_{k} \end{bmatrix},
\end{equation}
\begin{equation}
    J_r(\mathbf{q}):= \begin{bmatrix} -\mathbf{\ez}(\mathbf{q})^\times & 0 \\ 0 & I \end{bmatrix} J(\mathbf{q}).
\end{equation}
The next joint position is simply the current joint position incremented by $\dq$.  Since the Jacobian mapping is only approximate for finite increments, we optimize the increment size by writing $\mathbf{q}_k = \mathbf{q}_{k-1}+\alpha \delta q$ and choose $\alpha$ from a line search:
\[
\min_\alpha  \norm{\begin{bmatrix} \mathbf{\ez}(\mathbf{q}_{k})-\mathbf{n}^*_{k} \\ \mathbf{p}(\mathbf{q}_{k})-\mathbf{p}^*_{k} \end{bmatrix}}.
\]
The resulting sequence of joint angles  $\{\mathbf{q}_k\}_{k=0}^K$ now completely specifies the TCP position and orientation.


\subsubsection{Trajectory Traversal Speed Estimation}

\begin{figure}[h]
     \centering
     \begin{subfigure}[b]{0.22\textwidth}
         \centering
         \includegraphics[width=\textwidth]{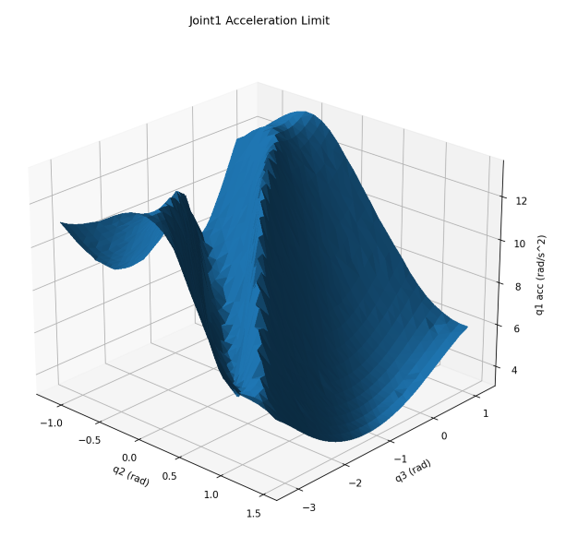}
         \caption{Joint1}
         \label{fig:j1_acc}
     \end{subfigure}
     \begin{subfigure}[b]{0.22\textwidth}
         \centering
         \includegraphics[width=\textwidth]{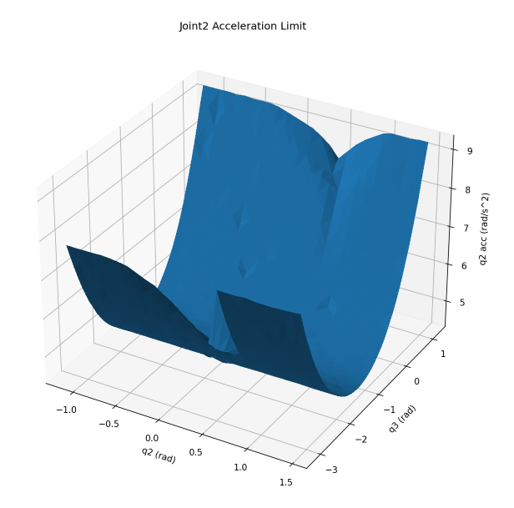}
         \caption{Joint2}
         \label{fig:j2_acc}
     \end{subfigure}
     \begin{subfigure}[b]{0.22\textwidth}
         \centering
         \includegraphics[width=\textwidth]{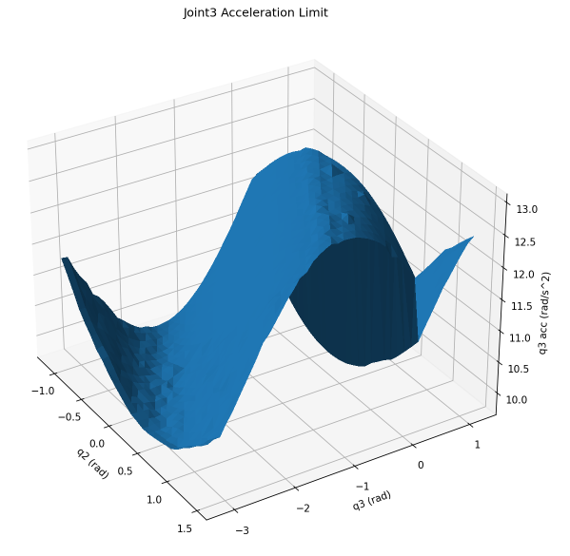}
         \caption{Joint3}
         \label{fig:j3_acc}
     \end{subfigure}
        \caption{Configuration Dependent Joint Acceleration Limit: Recorded maximum joint acceleration of joint 1,2,3 with respect to different joint 2 and 3 configuration.}
        \label{fig:j_acc}
\end{figure}

The robot TCP speed limit is determined by its joint velocity and acceleration limits.  Industrial robots provide joint velocity limits in the data sheets \cite{6640_manual}, but the acceleration limit is due to the torque limit and is therefore affected by the robot dynamics, load, and robot configuration. Since the dynamical models of industrial robots are typically unknown, we 
 establish a rough estimation by using vendor-provided dynamical simulation such as the ABB RobotStudio. 
 Since the arm inertia is most heavily influenced by the shoulder and elbow joints (joints 2 and 3), we assume that the spherical wrist joints (joint 4, 5, and 6) have constant joint acceleration limits, while joints 1, 2 and 3 acceleration limits depend on $(q_2,q_3)$. The recorded result is shown in Table~\ref{table:6640_limit} and Fig.~\ref{fig:j_acc}. 

\begin{table}[h!]
\centering
\begin{tabular}{||c c c c c c c||} 
 \hline
 Joint & 1 & 2 & 3 & 4 & 5 & 6 \\ [0.2ex] 
 \hline\hline
 $\qdot_{max}$ ($rad/s$)&   1.745 & 1.571 & 1.571 & 3.316 & 2.443 & 3.316   \\ 
 $\qddot_{max}$ ($rad/s^2$)&       *   & *     & *     & 42.5 &  36.8 & 50.5   \\[0.5ex] 
 \hline
\end{tabular}
\caption{ABB IRB6640 joint velocity and acceleration limits. * = configuration dependent, see Fig. \ref{fig:j_acc}.}
\label{table:6640_limit}
\end{table}

\def\qq{\mathbf{q}}
\def\qqdot{\mathbf{\dot q}}
\def\qqddot{\mathbf{\ddot q}}
\def\pp{\mathbf{p}}
The joint acceleration limits are stored as a look-up table dependent on $(q_2,q_3)$: $\qqddot_{\max}(q_2,q_3)$. For each robot joint angle point $\qq_k$ along the path (from Sec.~\ref{sec:toolorientation}), the maximum joint velocity limit is given by
\begin{equation}
    \qqdot_{\max_k} = \min\left(\qqdot_{\max},\qqdot_{k-1}+\Delta t_k \qqddot_{\max}(q_2,q_3) \right).
\end{equation}
For a specified (scalar) path speed $v_d$, $\pp_k = \pp_{k-1} + v_d \Delta t_k$.  Therefore,
\begin{equation}
    \Delta t_k = \frac{\pp_k-\pp_{k-1}}{v_d}.
\end{equation}
Hence the path velocity limit for point $k$ on the path is 
\begin{equation}
    v_k = \norm{J(\qq_k){\qq_{\max}}_k}.  
    \label{eq:vk}
\end{equation}


\subsubsection{Differential Evolution with Optimizing Parameters}

\begin{figure}[h]
     \centering
     \begin{subfigure}[b]{0.2\textwidth}
         \centering
         \includegraphics[width=\textwidth]{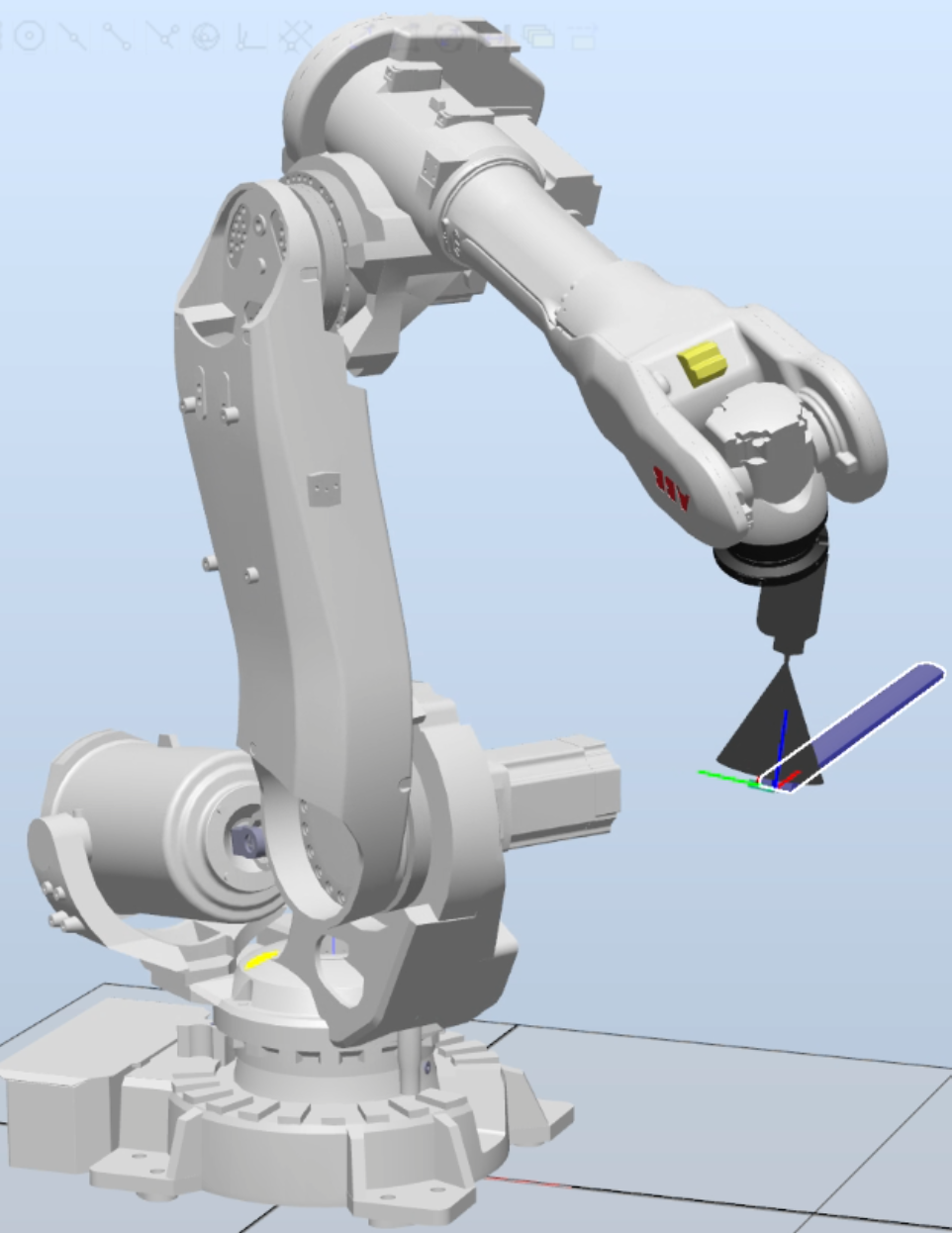}
         \caption{Curve1 Baseline Pose}
         \label{fig:curve1_baseline_pose}
     \end{subfigure}
     \begin{subfigure}[b]{0.2\textwidth}
         \centering
         \includegraphics[width=\textwidth]{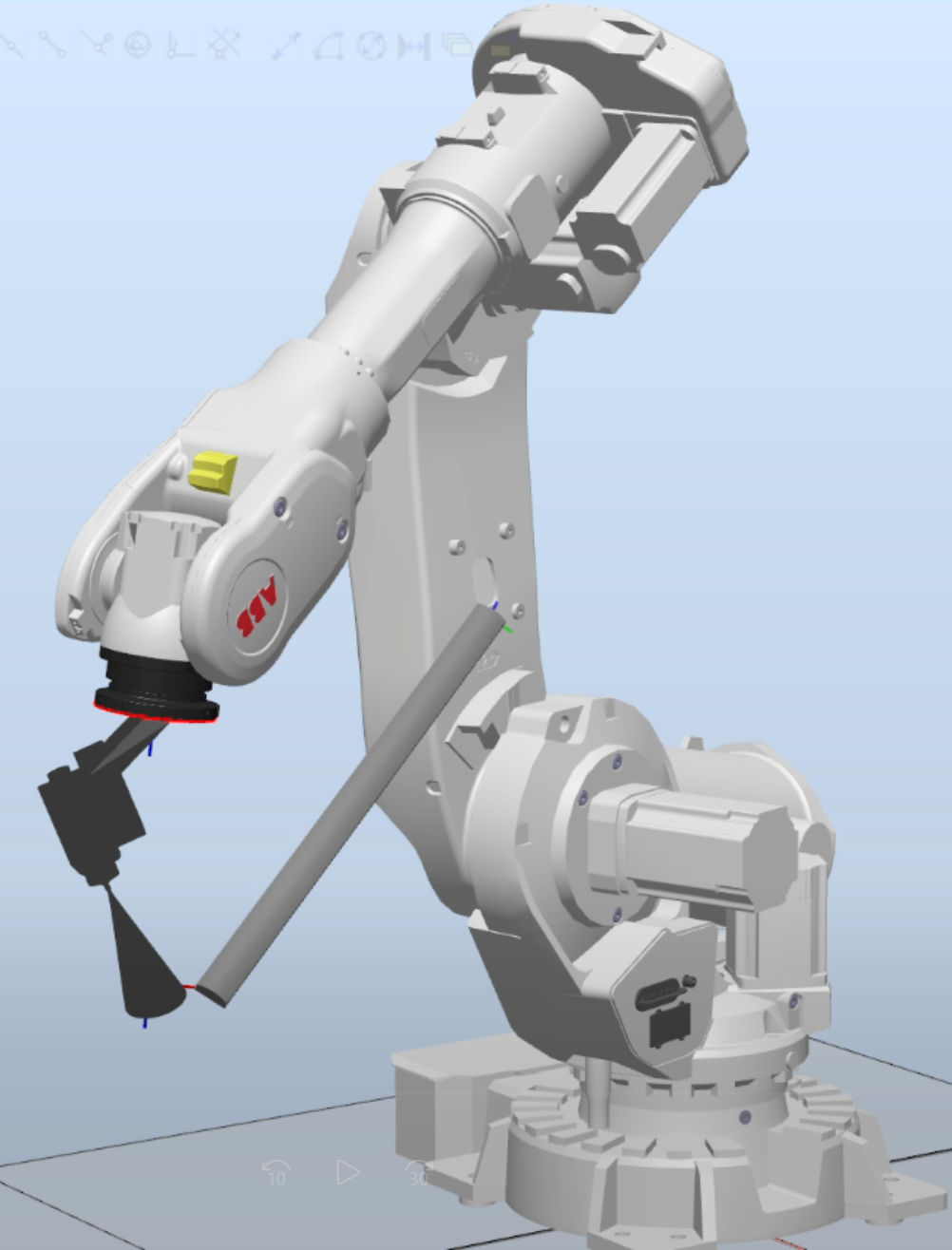}
         \caption{Curve1 Optimized Pose}
         \label{fig:curve1_opt_pose}
     \end{subfigure}
     \begin{subfigure}[b]{0.2\textwidth}
         \centering
         \includegraphics[width=\textwidth]{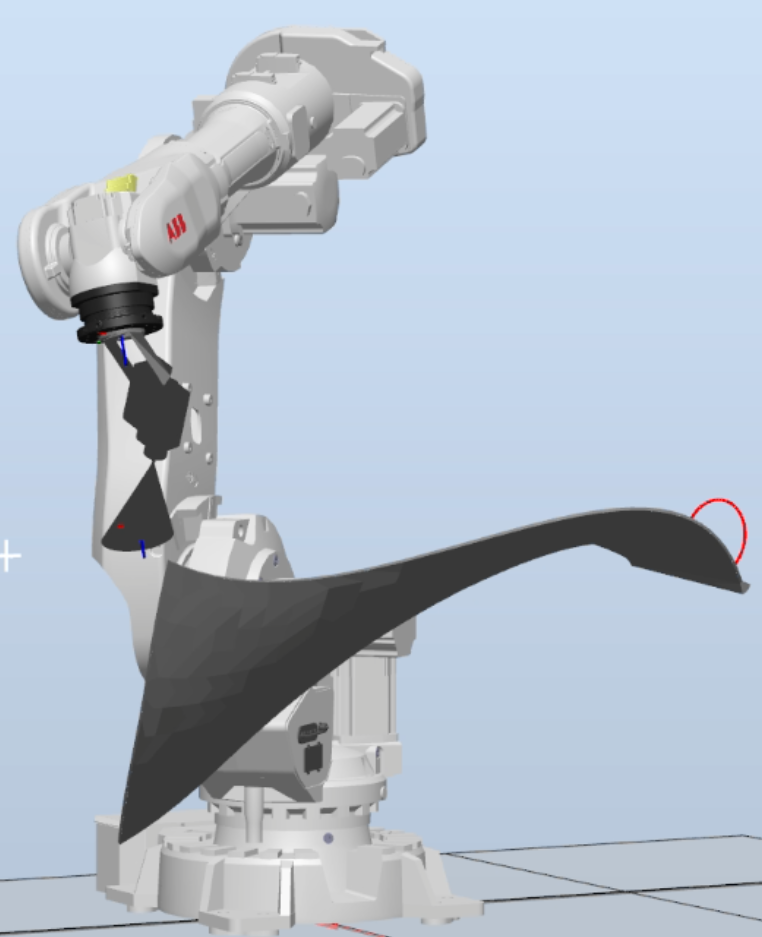}
         \caption{Curve2 Baseline Pose}
         \label{fig:curve2_baseline_pose}
     \end{subfigure}
     \begin{subfigure}[b]{0.2\textwidth}
         \centering
         \includegraphics[width=\textwidth]{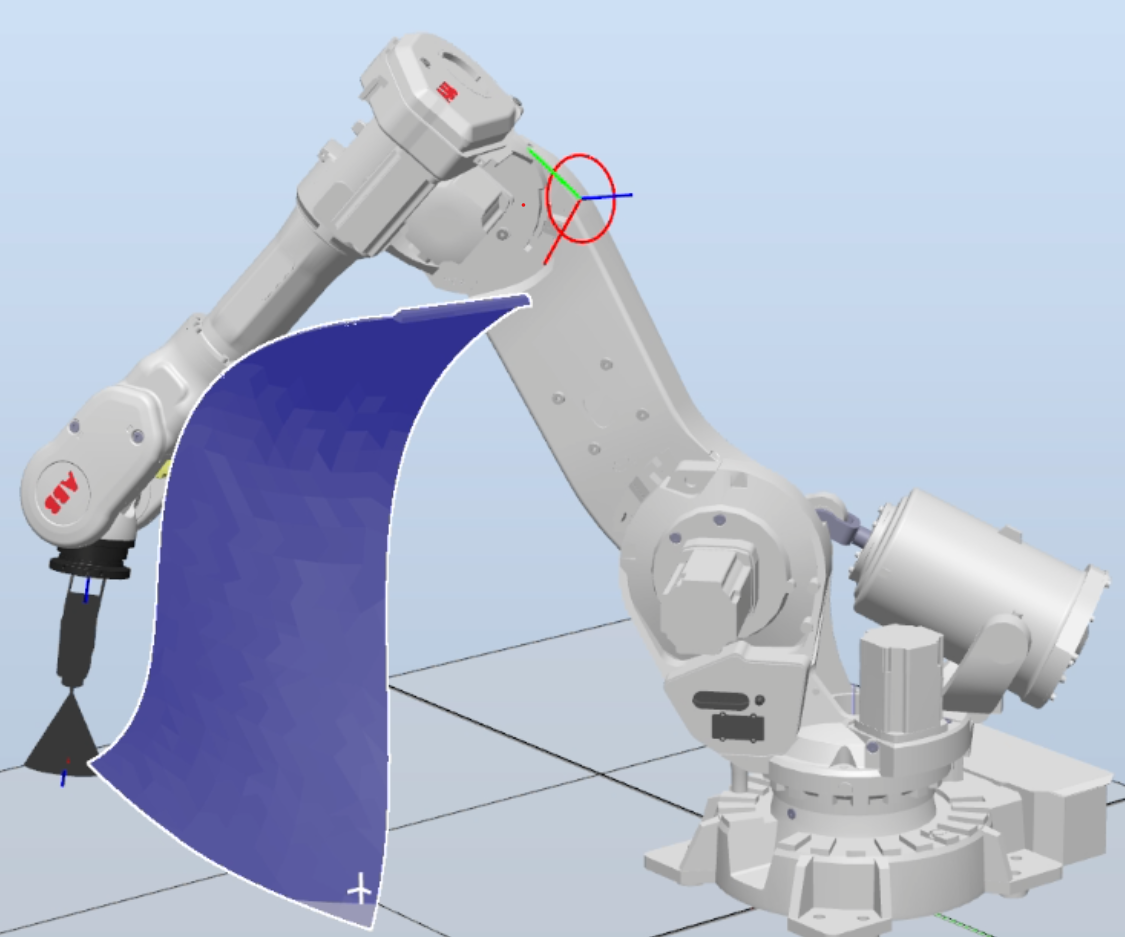}
         \caption{Curve2 Optimized Pose}
         \label{fig:curve2_opt_pose}
     \end{subfigure}
        \caption{Baseline curve pose (left) vs optimized curve pose (right) for Curve 1 and Curve 2 (described in Sec. \ref{sec:curves}.)}
        \label{fig:pose_opt}
\end{figure}

With the path speed limit parameterized by the curve pose and robot pose, we can optimize them to maximize the lowest speed limit:  ($\min(\{\mathbf{v}_k\}_{k=0}^K)$) could be formed as
\begin{equation}
\max_{\mathbf{p}_{curve},\bm{\beta}_{curve},\mathbf{q}_0} \min_k v_k
\label{eq:diff_evo}
\end{equation}
where $v_k$ is given by \eqref{eq:vk}. We use the differential evolution algorithm \cite{diff_evo}
to find the global optimizing solution for two example curves.  The optimized poses are as shown in Fig.~\ref{fig:pose_opt} versus the baseline (see Section~\ref{execution_baseline}).

\subsection{Motion Primitive Planning with Greedy Fitting}

\begin{figure}[h]
     \centering
     \begin{subfigure}[b]{0.23\textwidth}
         \centering
         \includegraphics[width=\textwidth]{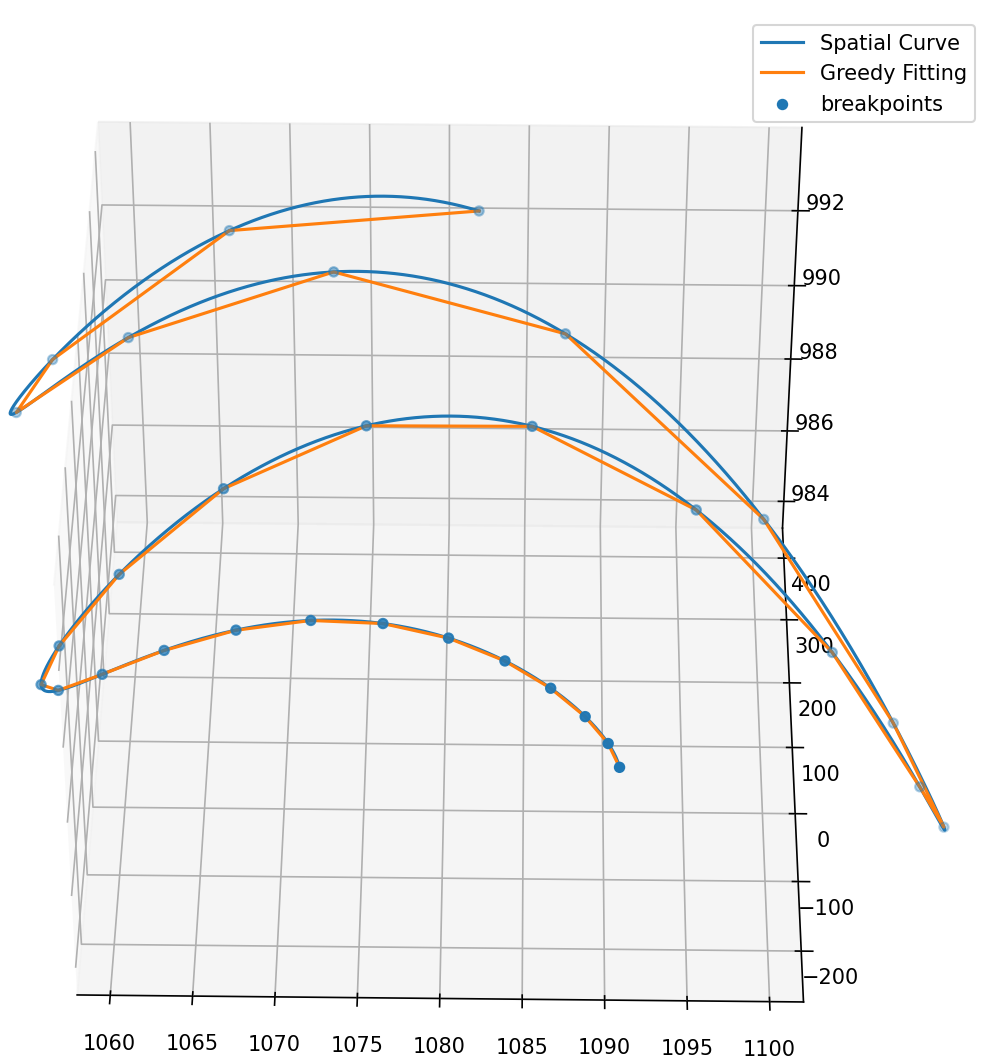}
         \caption{Equally spaced MoveL's}
         \label{fig:eq_movel}
     \end{subfigure}
     \begin{subfigure}[b]{0.23\textwidth}
         \centering
         \includegraphics[width=\textwidth]{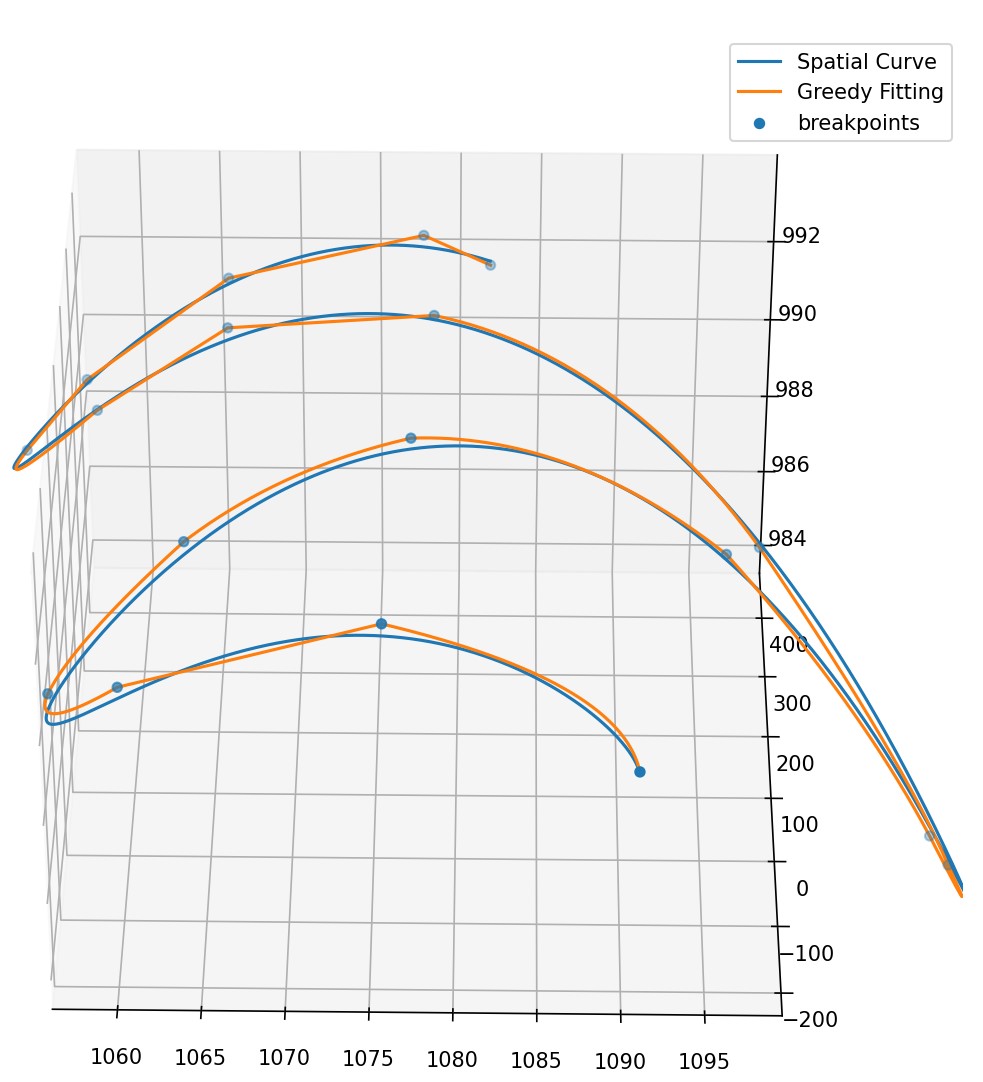}
         \caption{Greedy fitting}
         \label{fig:greedy}
     \end{subfigure}
        \caption{Curve fitting visualization.}
        \label{fig:fitting}
\end{figure}

Existing robot CAD/CAM software only use MoveL with a specified spatial curve. In order to take advantage of MoveC and MoveJ, we have developed a spatial curve (represented in joint space and Cartesian space) greedy fitting. The fitting process is performed sequentially along the trajectory, starting from the initial point and bisection search for the longest possible primitive type within the error threshold. For the initial segment, all regressions are unconstrained, while subsequent sequences are subject to continuity constraint (regression must pass through last segment endpoint). 
Fig.~\ref{fig:fitting} shows a visualization of 30 equally spaced MoveL's and Greedy Fitting with all three primitives on a test curve. While 30 linear segments will result in 1.47mm max position tracking error, greedy fitting only uses 17 segments to achieve a max position tracking error under 0.3mm.

\subsection{Waypoints Adjustment}
\label{waypoint_adjust}

The execution of the robot program based on the motion primitives would likely result in a trajectory that does not meet the performance specification. We can iteratively modify the waypoints to reduce the trajectory tracking error using a descent algorithm.   We use the following two algorithms:
\begin{itemize}
    \item Error Compensation: Push all waypoints in the error direction.
    \item Multi-peak Model Based Gradient Descent: Use an approximate trajectory 
    to calculate the numerical gradient and adjust waypoints based on the gradient descent. To simplify computation, we only apply to points at the error peaks.
\end{itemize}
Error compensation takes place in the first few iterations, aiming to bring down the tracking error at each waypoint while keeping the same primitive choices: $\begin{bmatrix} \mathbf{p}_{bp,j} \\ \bm{\beta}_{bp,j} \end{bmatrix}=\begin{bmatrix} \mathbf{p}_{bp,j} \\ \bm{\beta}_{bp,j} \end{bmatrix}+\gamma E(\begin{bmatrix} \mathbf{p}_{b,j} \\ \bm{\beta}_{bp,j} \end{bmatrix})$, where $E(\begin{bmatrix} \mathbf{p}_{bp,j} \\ \bm{\beta}_{bp,j} \end{bmatrix})$ is the error vectors at 
$j_{th}$ waypoint and $\gamma$ is the step size. 

Error compensation step only focuses the error at the waypoints, while the tracking error between waypoints may be unacceptably large.  We next apply a gradient descent method using an approximate numerical gradient to reduce the largest tracking errors.  For a given set of waypoints, we can efficiently generate a predicted joint trajectory using cubic spline interpolation at each waypoint. This predicted trajectory may then be used to find the numerical gradient by perturbing the waypoints and record the corresponding tracking error.  As the trajectory error is predominately affected by its neighboring waypoints, we only generate a tri-diagonal gradient, and only for the peak trajectory errors.

\section{Implementation and Evaluation}
\subsection{Test curves}
\label{sec:curves}


We use two representative curves. Curve 1, shown in Fig. \ref{fig:fitting}, is a multi-frequency sine curve on a parabolic surface. This curve is representative of a high curvature spatial curve. Curve 2, shown in Fig. \ref{fig:pose_opt}, is extracted from the leading edge of a generic fan blade model, which is a typical case for cold spraying applications in industry.



\subsection{Baseline}
\label{execution_baseline}
In consultation with industry experts, we establish a baseline performance based on the current practice.  The curve is placed in the middle of the robot workspace, away from singularities. The tool redundancy is resolved by choosing the tool $x$-axis along the curve.  The robot pose is chosen based on the largest manipulability measure. The motion primitives are based on equally spaced moveL segments.  We then search for the largest commanded path speed that satisfy both the path uniformity and trajectory tracking constraints as the baseline performance. 

\subsection{Simulation setup}
RobotStudio \cite{robotstudio} is an ABB robot dynamics simulator that provides close to real trajectory output. We use a Python interface \cite{abb_motion_program_exec} to RobotStudio virtual controller  to execute motion command directly. In order to achieve high constant speed, the commanded trajectory is extended by first and last command accordingly: MoveL and MoveJ is extended in Cartesian and joint space respectively, and MoveC is extend along the arc and linearly in angle-product orientation. For all motion commands, we start with a blending zone of 10 mm and gradually increase it until the speed profile converges such that the robot does not slow down around waypoints due to blending.  

\subsection{Experiment setup}
\begin{figure}[h]
    \centering
    \includegraphics[width=0.45\textwidth]{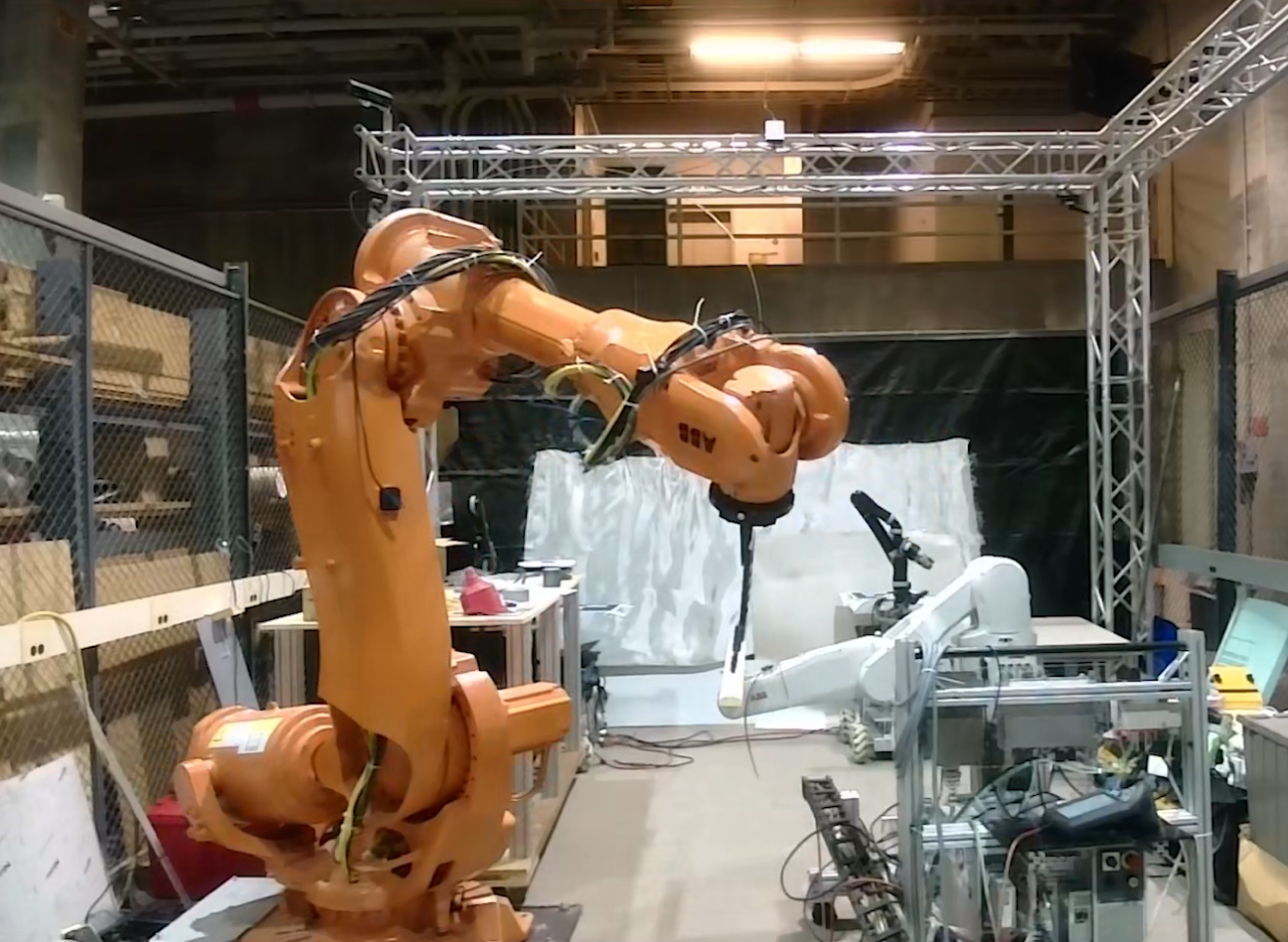}
    \caption{Experiment setup with an ABB6640 robot }
    \label{fig:setup}
\end{figure}

Fig.~\ref{fig:setup} shows our experiment setup, with the ABB6640 robot holding a mock spray gun. Joint trajectory $\{\mathbf{q}_{exe,i}\}_{i=1}^N$ is recorded through robot controller at 250Hz. We calculate the TCP position and orientation using forward kinematics of the with joint reading. To calculate the position and orientation tracking errrors, for each recorded joint position point $\mathbf{q}_{exe,i}$ we compute
\begin{align}
\mathbf{p}_{err,i} &:=\min_k \norm{\mathbf{p}(\mathbf{q}_{exe,i})-\mathbf{p}_k^*},\\
\mathbf{n}_{err,i} &:= \min_k \arccos \left ( \frac{\mathbf{e}^T_z(\mathbf{q}_{exe,i})\mathbf{n}_k^*}{\norm{\mathbf{e}_z(\mathbf{q}_{exe,i})}\norm{\mathbf{n}_k^*}}\right ).
\end{align}
The maximum position and orientation tracking errors are found by maximizing both terms above over $i$.


According to  the robot manual, the path repeatability is up to 1.06mm \cite{6640_manual}. To address this issue, for each set of primitive command, we run the robot 5 times and take the interpolated average of 5 recorded trajectories as the execution trajectory. 


\subsection{Results}

For baseline execution, in order to satisfy tracking accuracy requirement, we used bisection search on commanded speed until the accuracy requirement is met. For optimized trajectory, we run the post execution optimization to bring the tracking error until it converges. If it does not meet the requirement, we further slow down the commanded speed until the requirement is met.

\begin{table}[h!]
\centering
\begin{tabular}{||c c c c c||} 
 \hline
 Curve1 & $max(\mathbf{p}_{err})$ & $max(\mathbf{n}_{err})$ & $\mu(v)$ & $\sigma(v)$ \\ [0.2ex] 
 & mm & $^{\circ}$ & mm/s & mm/s \\
 \hline\hline
 Baseline (sim)&            0.49 & 0.18 & 124.01 & 1.02     \\ 
 Baseline (real)&           0.46 & 0.21 & 103.11 & 0.85     \\
 Bp Adjustment (sim)&       0.16 & 0.57 & 251.29 & 4.77     \\   
 Bp Adjustment (real)&      0.38 & 0.56 & 248.77 & 4.33     \\
 Optimized (sim)&           0.16 & 0.35 & 401.23 & 3.55     \\ 
 Optimized (real)&          0.39 & 0.48 & 395.86 & 8.06     \\[0.5ex] 
 \hline
\end{tabular}
\begin{tabular}{||c c c c c||} 
 \hline
 Curve2 & $max(\mathbf{p}_{err})$ & $max(\mathbf{n}_{err})$ & $\mu(v)$ & $\sigma(v)$ \\ [0.2ex] 
 & mm & $^{\circ}$ & mm/s & mm/s \\
 \hline\hline
 Baseline (sim)&            0.42 & 0.26 & 406.10    & 0.92     \\ 
 Baseline (real)&           0.47 & 0.28 & 299.97    & 2.36     \\
 Bp Adjustment (sim)&       0.24 & 0.33 & 1089.64   & 51.88    \\
 Bp Adjustment (real)&      0.42 & 0.41 & 973.12    & 13.87    \\
 Optimized (sim)&           0.44 & 0.67 & 1182.36   & 9.62     \\ 
 Optimized (real)&          0.50 & 1.37 & 1101      & 13.58    \\[0.5ex] 
 \hline
\end{tabular}
\caption{Execution Comparison for RobotStudio Simulation and Real Robot Experiments.}
\label{table:results}
\end{table}

From the baseline results, we can see that without any optimization, it is necessary to slow down in order to achieve higher accuracy. Simply adjusting waypoints though post execution optimization could help with increasing traversal speed and minimizing the tracking error. However, we can increase the traversal speed further by optimizing the all redundancies with robot joint velocity and acceleration constraints. The final optimized results show 222\% (sim) 283\% (real) speed increase for Curve 1 and 191\% (sim) and  267\% (real) speed increase for Curve 2 while keeping the tracking accuracy within the requirement. 

\subsection{Implementation for a FANUC Robot}

The same algorithm has also been applied to FANUC M710iC-70 robot using the FANUC robot simulation program RoboGuide. 
The result is showed in Table \ref{table:results_fanuc}. The speed increased by 323\% for curve1 and 265\% for curve2 while keeping tracking accuracy within the requirement. This shows that the framework successfully generalized to another robot model without fine tuning of the parameters.

\begin{figure}[h]
     \centering
     \begin{subfigure}[b]{0.233\textwidth}
         \centering
         \includegraphics[width=\textwidth]{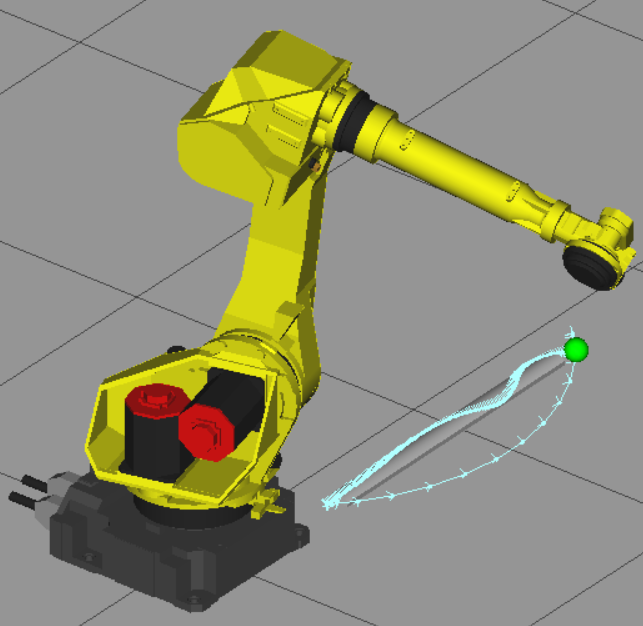}
         \caption{Curve 1}
     \end{subfigure}
     \begin{subfigure}[b]{0.22\textwidth}
         \centering
         \includegraphics[width=\textwidth]{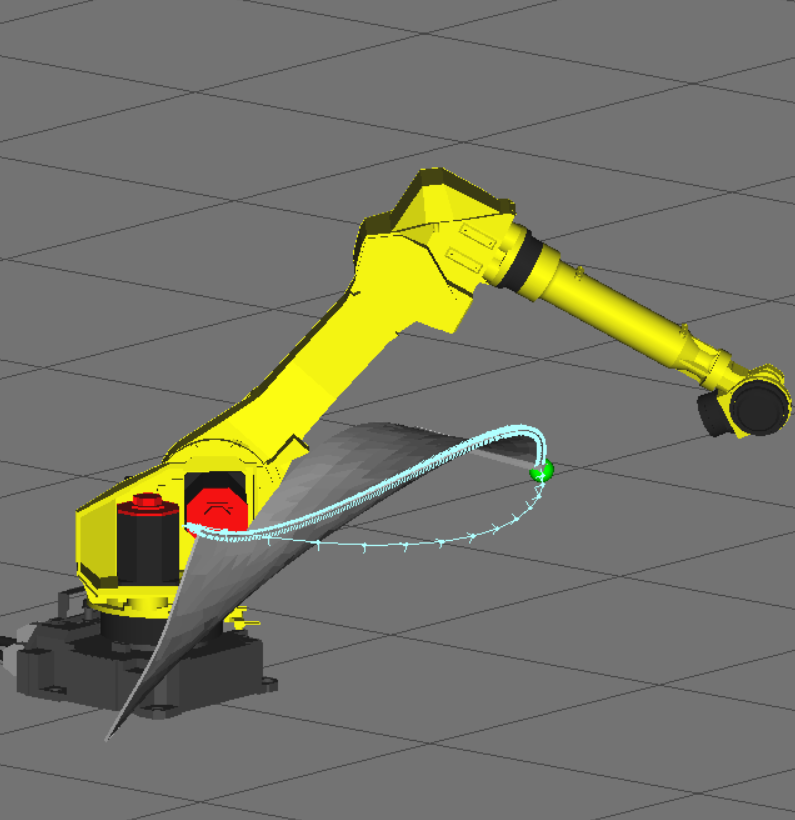}
         \caption{Curve 2}
     \end{subfigure}
     \caption{FANUC M710iC-70 robot in Roboguide.}
     \label{fig:fanuc_roboguide}
\end{figure} 

\begin{table}[h!]
\centering
\begin{tabular}{||c c c c c||} 
 \hline
 Curve1 & $max(\mathbf{p}_{err})$ & $max(\mathbf{n}_{err})$ & $\mu(v)$ & $\sigma(v)$ \\ [0.2ex] 
 & mm & $^{\circ}$ & mm/s & mm/s \\
 \hline\hline
 Baseline&            0.48 & 1.93 & 98.78 & 1.02     \\
 Bp Adjustment&       0.25 & 1.82 & 319.31 & 7.77     \\[0.5ex] 
 \hline
\end{tabular}
\begin{tabular}{||c c c c c||} 
 \hline
 Curve2 & $max(\mathbf{p}_{err})$ & $max(\mathbf{n}_{err})$ & $\mu(v)$ & $\sigma(v)$ \\ [0.2ex] 
 & mm & $^{\circ}$ & mm/s & mm/s \\
 \hline\hline
 Baseline&            0.43 & 0.92 & 274.61    & 2.48     \\ 
 Bp Adjustment&       0.38 & 0.75 & 728.45   & 10.83    \\[0.5ex] 
 \hline
\end{tabular}
\caption{Execution Comparison in RoboGuide}
\label{table:results_fanuc}
\end{table}

\section{Conclusion and Future Work}
We presented a complete procedure to generate motion primitive commands for industrial robots for tracking high curvature spatial curves with high speed and high accuracy. Our current experiment setup assumes perfect knowledge of the forward kinematics. In the future, we will consider a setup where the TCP position and orientation can be measured with a high precision scanner. Further, we will consider optimization of motion primitives in a dual-arm  setup where the performance is based on the relative TCP frames. 


\section*{ACKNOWLEDGMENT}
This research was supported by the ARM Institute project ARM-TEC-21-02-F-19. The authors would like to thank Santiago Paternain and Jonathan Fried for helpful discussion of the project. 

\bibliographystyle{IEEEtran}
\bibliography{bib}

\end{document}